\def\year{2022}\relax
\documentclass[letterpaper]{article} 
\usepackage{aaai22}  
\usepackage{times}  
\usepackage{helvet}  
\usepackage{courier}  
\usepackage[hyphens]{url}  
\usepackage{graphicx} 
\usepackage{natbib}  
\usepackage{caption} 
\DeclareCaptionStyle{ruled}{labelfont=normalfont,labelsep=colon,strut=off} 
\usepackage{algorithm}
\usepackage{algorithmic}
\usepackage{csquotes}

\usepackage{xcolor}

\usepackage{siunitx}

\usepackage{newfloat}
\usepackage{listings}
\floatstyle{ruled}
\newfloat{listing}{tb}{lst}{}
\floatname{listing}{Listing}

\usepackage{amsmath}
\usepackage{booktabs}

\usepackage{adjustbox}
\usepackage{makecell}

\title{Contrastive Explanations for Comparing Preferences of Reinforcement Learning Agents}
\author{
    Jasmina Gajcin, \textsuperscript{\rm 1}\textsuperscript{\rm \footnote{Work mostly done during an internship at IBM Ireland.}}
    Rahul Nair, \textsuperscript{\rm 2}
    Tejaswini Pedapati, \textsuperscript{\rm 2}
    Radu Marinescu, \textsuperscript{\rm 2}
    Elizabeth Daly, \textsuperscript{\rm 2}
    Ivana Dusparic \textsuperscript{\rm 1}
}
\affiliations{
    \textsuperscript{\rm 1} Trinity College Dublin \\
    \textsuperscript{\rm 2} IBM Ireland
   
    gajcinj@tcd.ie, \{rahul.nair, radu.marinescu, 	elizabeth.daly\}@ie.ibm.com, tejaswinip@us.ibm.com, 	ivana.dusparic@scss.tcd.ie 

}

\begin{document}

\maketitle

\begin{abstract}

In complex tasks where reward function is not straightforward and consists of a set of objectives, multiple reinforcement learning (RL) policies that perform task adequately, but employ different strategies can be trained by adjusting the impact of individual objectives on the reward function. Understanding the differences in strategies between policies is necessary to enable users to choose between offered policies, and can help developers understand different behaviors that emerge from various reward functions and training hyperparameters in RL systems. In this work we compare behavior of two policies trained on the same task, but with different preferences in objectives. We propose a method for distinguishing between differences in behavior that stem from different abilities from those that are a consequence of opposing preferences of two RL agents. Furthermore, we use only data on preference-based differences in order to generate contrasting explanations about agents' preferences. Finally, we test and evaluate our approach on an autonomous driving task and compare the behavior of a safety-oriented policy and one that prefers speed.

\end{abstract}

\section{Introduction}
\label{Introduction}

During the last decade, deep reinforcement learning algorithms (DRL) have shown notable results in a variety of applications, ranging from games to autonomous vehicles \cite{drl}. However, DRL algorithms rely on neural networks to represent agent's policy, making their decisions difficult to understand and interpret. This lack of transparency stands in the way of wider adoption of DRL methods in high-risk areas, such as healthcare or finance. Additionally, enabling agents to explain their decision-making process to humans is necessary to facilitate trust and collaboration between RL agent and user.

To address this issue, various approaches for interpreting behavior of DRL agents have been proposed in the recent years. Depending on their scope, methods for explaining a DRL system can either be local or global \cite{techniques-xai}. Local methods interpret a single decision of the RL model \cite{atari, fukuchi}, while global approaches explain policy behavior as a whole \cite{highlights, hayes}. Although these methods have shown to increase human understanding of agent's decision-making process, they have been mostly limited to explaining a single RL policy. However, comparing and
interpreting differences in policy behavior is necessary in situations where the user is confronted with a choice between alternative policies. Additionally, enabling developers to discover differences between policies could help them debug imperfect models by locating situations where they differ from a known expert model. Finally, comparing policies corresponds with the human tendency to prefer contrasting explanations which analyze differences in alternative scenarios \cite{miller}.

In this work we distinguish between two ways in which behaviors of two RL policies can differ: \textit{ability-based} and \textit{preference-based}. If one policy is trained to a higher standard and exhibits generally superior behavior compared to the other one, then differences between the two are ability-based. On the other hand, if both policies perform the task competently, but follow different strategies based on their individual inclinations then we consider their differences to be preference-based. Current methods for comparing RL policies assume that differences between policies are ability-based and completely disregard the notion of preference \cite{interestingness, disagreement-based-summaries}. However, in complex domains where defining the optimal behavior is not straightforward, it is possible to train multiple RL policies that perform the task adequately, but rely on different strategies. Interpreting differences between those policies is necessary from the perspective of personalisation -- if user needs to choose a policy that suits them best from a set of offered alternatives, understanding differences between them is crucial for making an informed decision. Furthermore, recognizing differences between multiple policies could aid developers in understanding the different behaviors that stem from various reward functions and hyperparameter combinations \cite{disagreement-based-summaries}. 

In RL, multiple different strategies can be the result of different reward functions in a task where the objective cannot be precisely defined, and certain trade-offs between goals have to be made \cite{enabling-robots-to-communicate}. For example, consider the problem of training a RL policy for driving an autonomous vehicle. Defining the reward function for this task is not straightforward and can involve multiple objectives that need to be satisfied. Consequently, multiple capable policies could be trained on different reward functions that slightly favor a specific objective over the others. These policies will achieve high average reward according to their individual reward function, but due to their opposing preferences in terms of objectives they may also exhibit different strategies for performing the task. Specifically, a policy which was trained on the reward function that severely penalizes close contact with another car will develop a safety-oriented strategy, and prefer slower driving and keeping distance from other vehicles. On the other hand, policy that was penalized for not arriving to the destination on time  will likely prefer faster driving, and will have a more relaxed understanding of safety concerns. If a user is given a choice between these two policies it is necessary that the they recognize the differences between their strategies to choose the one best fitting for them. 

In this work, we focus on comparing two policies trained on the same task and interpret their preference-based differences. We choose the global approach to explainability, and attempt to discover the differences in overall behavior of two RL agents as opposed to analysing their differences in a single state. Our approach attempts to uncover differences between two policies by analysing situations where policies disagree on the best strategy to follow. We propose an algorithm for distinguishing between disagreements that stem from difference in ability between two agents and those that arise from their opposing preferences. We then analyse only the preference-based disagreements in order to extract conditions in terms of state features that specific agents favour and generate global explanations contrasting agents' behavior. We test and evaluate our approach in an autonomous driving environment, where agent's task is to merge into another lane currently occupied by a non-autonomous vehicle, and we compare policies of a safety-oriented vs  speed-oriented agent. 

Our contributions are as follows:
\begin{itemize}
    \item We propose a method for distinguishing between ability-based and preference-based differences in behavior between two RL agents.
    \item We present a method for generating contrasting explanations based on the contrast in preferred state feature values between two policies.
    \item We test and evaluate our approach in an autonomous driving environment.
\end{itemize}

\section{Related work}
In the recent years various methods for explaining either one decision or the entire behavior of RL system have been proposed \cite{xrl}.
However, despite the fact that research shows humans tend to seek contrasting explanations when reasoning about an event \cite{miller}, there is still limited work in the field of comparing alternative behaviors of reinforcement learning systems.

\citet{causal-lens, distal} approached the problem of explainability from a causal perspective and proposed a method for generating local explanations that contrast alternative actions in a specific state. The approach however requires a hand-crafted causal model of the environment, which may be difficult to obtain and requires expert knowledge. Additionally, authors focus on generating local explanations, while we interpret and compare global behavior of agents.

Summarisation methods, one of the most notable global approaches for condensing and explaining the agent behavior have also been used to compare multiple RL policies. \citet{interestingness} generated contrasting summaries of agents' behavior in order to highlight the differences in their capabilities. Similar to our work, \citet{disagreement-based-summaries} used the notion of disagreement between policies to detect and analyse situations where two policies pick different actions, but opted for explaining policies' differences through contrasting summaries. However, both approaches focus only on extracting discrepancies between agents' abilities, and disregard the potential difference in their strategies. Additionally, applicability of summarisation methods is limited to tasks with visual input. 

Most relevant to our work, \citet{contrastive-exp} compare the outcomes of following different policies from a specific state to justify agent's choice of action. However, their work focuses only on local explanations, and requires manual encodings of states and outcomes. In contrast, in our work we aim to provide global comparisons of policies and do not rely on hand-crafted interpretable features. 

\section{Preference-based contrastive explanations}
\label{preference-based-explanations}

In this section we propose a set of conditions for distinguishing between ability and preference-based differences between two policies and offer a method for extracting explanations that highlight feature values that specific agents favour. Throughout this section we assume oracle access to two policies $\pi_A$ and $\pi_B$, their $Q$ action-value functions and the transition function of the environment $T$. Our approach consists of three steps presented in this section. Firstly, policies are unrolled in the environment to collect data on situations where two policies disagree on the best course of action (Section \ref{disagreement-data}). Afterwards, collected data is filtered so that only data illustrating preference-based differences between the policies is obtained (Section \ref{filter}). Finally, we analyse and compare preference-based disagreement data from both policies to extract explanations that indicate which conditions agents prefer to end up in (Section \ref{generating-exp}). 

\subsection{Disagreement data}
\label{disagreement-data}

We adopt the definition of \textit{disagreement state} from \citet{disagreement-based-summaries} and consider two policies to disagree in a state if they do not choose the same action in that state. 

With that in mind, we collect three different types of disagreement data from policies' interaction with the environment.
We follow the method for gathering disagreement data presented in \citet{disagreement-based-summaries} which assumes unrolling policy $\pi_A$ in the environment and at every step comparing decisions of $\pi_A$ and $\pi_B$ until a disagreement is reached, then following both policies separately for a set number of steps $k$, and finally returning control to $\pi_A$.

Specifically, we start by executing policy $\pi_A$ in the environment. In each state $s$ that $\pi_A$ encounters we compare the decisions of both policies and record those states in which policies choose different actions: 

\noindent
\textbf{Definition} (Disagreement states)
\textit{Given two policies $\pi_A$ and $\pi_B$, $s_d$ is a disagreement state if}:
\begin{equation}
     \pi_A(s_d) \ne \pi_B(s_d) 
\end{equation}

After encountering a disagreement state $s_d$, we also unroll both policies for a set number of steps $k$ starting from $s_d$ and record the resulting pair of trajectories:

\noindent
\textbf{Definition} (Disagreement trajectories):
\textit{Given two policies $\pi_A$ and $\pi_B$ and a disagreement state $s^t_d$, a pair of disagreement trajectories is a tuple $(\mathcal{T}_{\pi_A}(s^t_d)$, $\mathcal{T}_{\pi_B}(s^t_d))$ where}:

\begin{equation}
    \begin{aligned}
        \mathcal{T}_{\pi_A}(s^t_d) = \{s_d^t, \dots s^{t+k}_d\} \\
        \mathcal{T}_{\pi_B}(s^t_d) = \{s^t_d, \dots s^{t+k}_d\} 
    \end{aligned}
\end{equation}

Finally, upon collecting disagreement trajectories, we also record the last states in each trajectory pair:

\noindent
\textbf{Definition} (Disagreement outcomes):
\textit{Given two policies $\pi_A$ and $\pi_B$ and a pair of disagreement trajectories $(\mathcal{T}_{\pi_A}(s_d), \mathcal{T}_{\pi_B}(s_d))$, where $\mathcal{T}_{\pi_A}(s_d) = \{s^{i}_d\}_{i \in [1...k]}$ and $\mathcal{T}_{\pi_B}(s_d) = \{s^{i}_d\}_{i \in [1...k]}$, pair of disagreement outcomes is a tuple $(o_A(s_d), o_B(s_d))$ where}:
\begin{equation}
    \begin{aligned}
    o_A(s_d) = \mathcal{T}_{\pi_A}[k] \\
    o_B(s_d) = \mathcal{T}_{\pi_B}[k]
    \end{aligned}
\end{equation}

In other words, an outcome is a state in which agent ends up after following its policy for a set number of steps from a disagreement state.
After individually unrolling two policies from disagreement state $s_d$ for $k$ steps, control is returned to policy $\pi_A$ which continues to progress in the environment, until a new disagreement state or episode terminates. Entire collection process is repeated for $n$ episodes. The approach is further detailed in Algorithm \ref{collecting-data}.

Throughout this section we use the term \textit{disagreement} to denote a tuple $(s_d, \mathcal{T}_{\pi_A}(s_d), \mathcal{T}_{\pi_B}(s_d), o_A(s_d), o_B(s_d))$ where $s_d$ is a disagreement state, $\mathcal{T}_{\pi_A}(s_d)$ and $\mathcal{T}_{\pi_B}(s_d)$ disagreement trajectories starting in $s_d$, and $o_A(s_d)$ and $o_B(s_d)$ their outcomes. Output from this section of the approach is a set of collected disagreements $D(\pi_A, \pi_B)$.

\begin{algorithm}[t]
    \begin{algorithmic}[1]
        \STATE Input: $\pi_A$, $\pi_B$, $T$
        \STATE Parameters: $n$, $k$
        \STATE Output: $D(\pi_A, \pi_B)$
        \STATE $D(\pi_A, \pi_B) = \{\}$
        \FOR{episode $i \in [0, n]$}{
            \STATE $s$ = $T.next\_state()$ 
            \WHILE{episode not done}{
                \STATE $a_A = \pi_A(s)$, $a_B = \pi_B(s)$
                \IF{$a_A \ne a_B$}{
                    \STATE $\mathcal{T}_A = \{\}$, $\mathcal{T}_B = \{\}$
                    \STATE $s_j = s$, $a_j = a_A$
                    \FOR{$j \in [0, k]$}{
                        \STATE $\mathcal{T}_A \mathrel{{+}{=}} (s_j, a_j)$
                        \STATE $s_j = T.act(a_j)$
                        \STATE $a_j = \pi_A(s_j)$
                    }\ENDFOR
                    \STATE $T.reset(s)$ 
                    \STATE $s_l = s$, $a_l = a_B$
                    \FOR{$l \in [0, k]$}{
                        \STATE $\mathcal{T}_B \mathrel{{+}{=}} (s_l, a_l)$
                            \STATE $s_l = T.act(a_l)$
                            \STATE $a_l = \pi_B(s_l)$
                    }\ENDFOR
                    \STATE $T.reset(s_j)$
                    \STATE $D(\pi_A, \pi_B) \mathrel{{+}{=}} (s, \mathcal{T}_A, \mathcal{T}_B, o_A, o_B)$
                }\ENDIF
                \STATE $s = T.act(a_A)$
            }\ENDWHILE
        }\ENDFOR
    \end{algorithmic}
    \caption{Collecting disagreement data}
    \label{collecting-data}
\end{algorithm}

\subsection{Ability vs. preference-based disagreement}
\label{filter}

In order to generate preference-based explanations using the gathered disagreement data, we need to distinguish between disagreement that comes from different abilities of two agents and that which is a consequence of different preferences. Specifically, we aim to a detect particular type of disagreement that can be representative of difference in preferences between two policies. Intuitively, we would like to select only those disagreements $(s_d, \mathcal{T}_{\pi_A}(s_d), \mathcal{T}_{\pi_B}(s_d), o_A(s_d), o_B(s_d))$ where both agents see the same potential in the state $s_d$ and  fulfil that potential to the same extent over the next $k$ steps, but disagree strongly on the course of action in $s_d$. In other words, we select disagreements where agents feel similarly optimistic in state $s_d$ and policy $\pi_A$ feels similarly as satisfied being in $o_A$ as $\pi_B$ is with reaching $o_B$, while both agents have high confidence in their chosen path. This would indicate that agents estimate and realize the same potential in the environment, but strongly disagree on their preferred way to do so. 

To precisely define these conditions, we need to introduce a metric for measuring how strongly two policies disagree in a specific state. For this purpose we define \textit{state importance} as follows:

\noindent
\textbf{Definition} (State importance) \textit{Given two policies $\pi_A$ and $\pi_B$, a disagreement state $s_d$, and if $Q_A(s_d)$ and $Q_B(s_d)$ are vectors of Q action-values in state $s_d$ according to policies $\pi_A$ and $\pi_B$ respectively, the importance of $s_d$ with regards to policies $\pi_A$ and $\pi_B$ can be defined as:}

\begin{equation}
\resizebox{0.9\hsize}{!}{$\mathcal{I}(s_d | \pi_A, \pi_B) = \frac{\max(softmax(Q_A(s_d))) + \max(softmax(Q_B(s_d)))}{2}$}
\end{equation}

We compute the softmax over a vector of Q action-values in order to emphasise the contrast between the first-ranked action and the others, and select the maximum value to represent how sure the policy is in its decision.

Furthermore, in order to quantify the potential that policy sees in a specific state we employ the idea of a state-value function. Since we do not assume direct access to state-value functions of policies, we simulate them with the help of the available Q action-value functions:

\begin{equation}
    V_{\pi}(s) = \max_{a \in \mathcal{A}} (Q_{\pi}(s, a))
\end{equation}

Before using them for estimating state-value function, Q action-values are normalized to $\begin{bmatrix} 0, 1 \end{bmatrix}$ range, so that they can be compared between different agents.

Finally, we formalize the conditions for considering disagreement $d = (s_d, \mathcal{T}_{\pi_A}(s_d), \mathcal{T}_{\pi_B}(s_d), o_A(s_d), o_B(s_d))$ to be preference-based:

\noindent
\textbf{Definition} (Preference-based disagreement) \textit{Given two policies $\pi_A$ and $\pi_B$ and disagreement between them $d = (s_d, \mathcal{T}_A(s_d), \mathcal{T}_B(s_d), o_A(s_d), o_B(s_d))$, disagreement is considered to be preference-based if the following conditions are fulfilled:}
\begin{enumerate}
    \item  \textit{Both policies are highly confident in their decision in the disagreement state $s_d$:}
    \begin{equation}
        \mathcal{I}(s_d| \pi_A, \pi_B) > \alpha
    \end{equation}
    \item  \textit{Both policies have similar evaluations of the disagreement state $s_d$:}
    \begin{equation}
        V_{\pi_A}(s_d) \approx V_{\pi_B}(s_d)
    \end{equation}
    \textit{To estimate this similarity we evaluate the expression:}
    \begin{equation}
        \lvert V_{\pi_A}(s_d) - V_{\pi_B}(s_d)\rvert < \beta
    \end{equation}
    \item \textit{After unrolling policies in the environment for $k$ steps from state $s_d$, both policies have similar evaluations of their outcomes:}
    \begin{equation}
        V_{\pi_A}(o_A) \approx V_{\pi_B}(o_B)
    \end{equation}
    \textit{To estimate this similarity we evaluate the expression:}
    \begin{equation}
        \lvert V_{\pi_A}(o_A) - V_{\pi_B}(o_B)\rvert < \gamma
    \end{equation}
\end{enumerate}
\textit{where $\alpha$, $\beta$ and $\gamma$ are threshold values}.

Selected values for threshold parameters $\alpha$, $\beta$ and $\gamma$ affect how selective the algorithm is. Some suitable values to use in practice would be: $0.8 \le \alpha \le 1.0$, $\beta \le 0.1$ and $ \gamma \le 0.1$. Decreasing $\alpha$ would 
result in including disagreements where policies are not confident in their choice and evaluate some alternative actions as similarly promising. These situations indicate that the disagreement is not a consequence of strong preferences of the agents. On the other hand increasing parameters $\beta$ or $\gamma$ would result in allowing more ability-based disagreements into the end result. For example, consider a situation where two agents encounter state $s_d$ which they evaluate similarly, but disagree strongly on the best course of action. If after unrolling these policies separately for a number of steps they arrive at different states, and one policy is far more satisfied with its outcome than the other, that indicates that the disagreement in $s_d$ was not a consequence of opposing preferences, but rather of inferior abilities of the second policy. 

Finally, we can use the defined measures to filter the set of disagreements to obtain only those that are preference-based:
\begin{equation}
    \begin{aligned}
    D_{p}(\pi_A, \pi_B) = \{d \in D(\pi_A, \pi_B) | 
     \mathcal{I}(s_d; \pi_A, \pi_B) > & \alpha, \\ 
     \lvert V_{\pi_A}(s_d) - V_{\pi_B}(s_d)\rvert < & \beta, \\
     \lvert V_{\pi_A}(o_A) - V_{\pi_B}(o_B)\rvert < & \gamma\}
    \end{aligned}
\end{equation}

Output from this stage is the set of preference-based disagreements $D_{p}(\pi_A, \pi_B)$.

\subsection{Generating contrastive explanations}
\label{generating-exp}

Data set $D_p(\pi_A, \pi_B)$ is rich with information on preference-based differences between $\pi_A$ and $\pi_B$. Therefore, there are multiple way to approach generating contrastive explanations using $D_p(\pi_A, \pi_B)$ -- we could exploit differences in disagreement trajectories or disagreement outcomes. In this work we choose the latter and focus on analysing and comparing outcomes of following the two policies to uncover which states agents prefer to reach. More specifically, we address the question: ``\textit{What conditions (in terms of state features) does agent $A$ prefer to end up in, compared to agent $B$?"}. To answer this question, we start by creating two sets of values $O^f_A$ and $O^f_B$ for each state feature $f$, storing values of feature $f$ in outcomes from policies $\pi_A$ and $\pi_B$ respectively:

\begin{equation}
    \begin{aligned}
        O^f_A = \{o_A[f] | o_A \in D_p(\pi_A, \pi_B)\}\\
        O^f_B = \{o_B[f] | o_B \in D_p(\pi_A, \pi_B)\}
    \end{aligned}
\end{equation}

Finally, we include in our explanations only those features for which there is a significant difference in distributions of $O_A$ and $O_B$. Since we deal with continuous state features in the example presented in Section \ref{experiments}, we use a paired T-test to assess the difference in the two distributions. If $f$ is not continuous, alternative appropriate statistical tests can be used to determine statistical relationship \cite{mcnemar, chi-square}. Provided that there is a significant difference in distribution of feature $f$ in outcomes of the two policies, we consider $f$ to be indicative of agent's preference and we use it in the explanations. For example, if there is a significant difference between $O_A[f]$ and $O_B[f]$ for some feature $f$ and mean value of $O_A[f]$ is larger than mean value of $O_B[f]$ then the explanation will include that $\pi_A$ prefers to end up in states where feature $f$ has larger values. Final explanation is generated by combining all feature-specific preferences using a natural-language template.
 
\section{Experiments}
\label{experiments}

\begin{figure}[t]
    \centering
    \includegraphics[width=\linewidth]{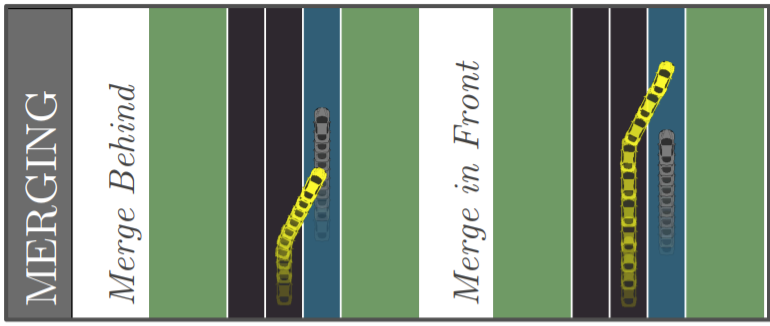}
    \caption{Autonomous driving environment for the merging task as presented in \cite{enabling-robots-to-communicate}}
    \label{merging}
\end{figure}

To evaluate the method proposed in Section \ref{preference-based-explanations}, we employ a simplified version of the merging task in autonomous driving presented in \citet{enabling-robots-to-communicate}. In this environment, the autonomous vehicle navigates a three-lane road. Agent begins the episode in the center lane, and is tasked with merging safely into the right lane, currently occupied by a non-autonomous vehicle. Episode  ends upon successful completion of the task, or if agent fails catastrophically by crashing into another car. Reward of $+1000$ is awarded for successfully merging in the right lane, while $-1000$ penalty is received for crashing into the non-autonomous vehicle. Additionally, driving off the road yields $-10$ penalty. 
There are two suitable ways to approach this task. Agent can either employ a safety-oriented strategy and merge behind the non-autonomous vehicle, minimizing its chances of collision or it can speed up and merge in front of the other car, depending on its preference in the trade-off between speed and safety.

Agent's observation is a vector describing both the agent and the non-autonomous vehicle. Specifically, agent observes location of its rear axle $(x, y)$, as well as its heading $h$, velocity $v$ and steering wheel angle $u$. Additionally, agent can observe the same features for the non-autonomous vehicle denoted by $x', y', h', v'$ and $u'$. At each step agent chooses between $5$ discrete actions -- agent can increase or decrease its speed by $10\%$, it can change its steering angle by $3$ degrees in any direction, or it can choose to alter nothing. Non-autonomous vehicle, however, drives straight ahead in the right lane with the same velocity throughout the episode. Action space is significantly simplified compared to one described in \cite{enabling-robots-to-communicate}, but it is still rich enough to enable the desired behavior. Additional environment parameters are available in Table \ref{env-params}. Finally, the reward function consists of multiple features that agent needs to optimize: 
\begin{itemize}
    \item \textit{Distance from the goal}:
    $
        \mathcal{D}^t_{g} = \max(0, |x_{goal} - x_t|) 
    $,
    where $x_{goal}$ is the x coordinate of the center of the right lane. 
    \item \textit{Distance to the other car}:
    $
        \mathcal{D}^t_{c} = e^{-\frac{\|p - p'\|}{2\sigma^2}}
    $,
    where $p_t = (x_t, y_t)$ is the location of the agent and $p' = (x_t', y_t')$ is the location of the non-autonomous car at time step $t$.
    \item \textit{Deviation from initial speed}:
    $
        \mathcal{D}^t_{v} = |v_t - v_0|
    $
    \item \textit{Deviation from initial heading}:
    $
        \mathcal{D}^t_{h} = |u_t - u_0|
    $
    \item \textit{Progress}:
    $
        \mathcal{D}^t_{p} = |y_t - y_{t-1}| 
    $
\end{itemize}

\begin{table}[t]
    \centering
    \caption{Environment parameters for the merging task.}
    \begin{adjustbox}{width=0.7\linewidth}
        \begin{tabular}{@{}cc@{}}
            \toprule
            Parameter description & Value \\ \midrule
            Number of lanes & 3 \\
            Lane width & 10 \\ 
            Car length & 10 \\
            Initial velocity (for both vehicles) & 15 \\
            Maximum velocity & 20 \\
            Steering angle range & $\begin{bmatrix} -30, 30 \end{bmatrix}$ \\
            \bottomrule
        \end{tabular}
    \end{adjustbox}
    \label{env-params}
\end{table}

All reward features are normalized to $\begin{bmatrix} 0, 1 \end{bmatrix}$ range. Ultimately, the reward function takes the form of a linear combination of features:

\begin{equation}
    \mathcal{R}(s_t, a_t) =
         \begin{bmatrix}
            \mathcal{D}^t_{g} &
            \mathcal{D}^t_{c} &
            \mathcal{D}^t_{v} &
            \mathcal{D}^t_{h} &
            \mathcal{D}^t_{p}
        \end{bmatrix}
        \theta^\mathsf{T}
\end{equation}

where each feature is weighed by a parameter from the vector $\theta = \begin{bmatrix} \theta_0, \theta_1, \theta_2, \theta_3, \theta_4 \end{bmatrix}$ to determine its importance in the overall objective. 

We start by training a baseline policy $\pi_{safe}$ with reward function parameters $\theta_{safe} = \begin{bmatrix} 5, 10, 20,  50, 0 \end{bmatrix}$. This policy is safety-oriented -- it prefers to keep a distance to the non-autonomous vehicle and chooses to slow down and merge behind it. All policies in this section are trained using DQN algorithm \cite{dqn}. 

Furthermore, to obtain policies with different strategies we vary the value of the parameter $\theta_4$ which affects how important progress is to the agent. Small values of this parameter indicate a safety-oriented policy which prefers to slow down and merge behind the non-autonomous vehicle, decreasing its chances of collision. This strategy is identical to that of $\pi_{safe}$. On the other hand, increasing the value of this parameter results in a more aggressive policy which values progress over keeping greater-than-necessary distance to the other vehicle. Such policy prefers to speed up and merge in front of the non-autonomous car. For each value $p \in \begin{bmatrix} 0, 1, 2, 3, 4 \end{bmatrix}$ we train $3$ different models $\pi^{p}_A, \pi^{p}_B$ and $\pi^{p}_{rand}$ using the reward function with parameters $\theta_p = \begin{bmatrix} 5, 10, 20, 50, p \end{bmatrix}$. In other words, we keep all other reward feature parameters same as in the baseline model, and change only the parameter corresponding to progress. Models $\pi^{p}_A$ and $\pi^{p}_B$ are trained for the same number of steps and achieve near-optimal performance on the task. To ensure we also generate a policy with inferior capabilities, $\pi^{p}_{rand}$ is trained for significantly less time, and does not fully learn the task. 
Training parameters are given in Table \ref{train-params}.

\begin{table}[t]
    \centering
    \caption{Training parameters for learning policies $\pi^p_A, \pi^p_B, \pi^p_{rand}, \pi_{safe}$ with DQN algorithm. Same parameters are used for all values of $p$.}
    \begin{adjustbox}{width=\linewidth}
        \begin{tabular}{@{}cccccccccc@{}}
            \toprule 
            Architecture & \multicolumn{4}{c}{Train time steps} & Learning rate & Batch size & Buffer size & Exploration fraction & Final exploration $\epsilon$ \\ \cmidrule(r){2-5}  
               &  $\pi_{safe}$ & $\pi^{p}_A$ & $\pi^{p}_B$ & $\pi^{p}_{rand}$ &  & \\ \midrule
             \makecell[l]{Linear(10, 256) \\ Linear (256, 1)} & $\num{e5}$ & $\num{e5}$ & $\num{e5}$ & $\num{e4}$ & $\num{6.3e-4}$ & $128$ & $\num{5e4}$ & $0.1$ & $0.01$\\
            \bottomrule
        \end{tabular}
    \end{adjustbox}
    \label{train-params}
\end{table}

\section{Evaluation}

We set up experiments to show that the method presented in Section \ref{preference-based-explanations} captures and explains only preference-based differences in behavior of the two agents. In other words, we do not want our method to detect difference in behavior when comparing two policies employing the same strategy, or two policies with significantly different capabilities. Therefore, we set up three different evaluation goals:

\begin{enumerate}
    \item The method detects differences between two policies with different preferences.
    \item The method does not detect differences between two policies with same preference.
    \item The method does not detect differences between two policies of significantly different capabilities.
\end{enumerate}

To test all three evaluation goals, we set up three different evaluation scenarios. To evaluate our method against the first goal, we compare the behavior of policies $\pi_{safe}$ and $\pi^{p}_A$ for each $p \in \begin{bmatrix} 0, 1, 2, 3, 4\end{bmatrix}$. Since we expect $\pi^p_A$ to exhibit more aggressive behavior compared to $\pi_{safe}$ for larger values of $p$, this scenario tests whether this difference in strategy will be captured by the proposed approach. Furthermore, to evaluate the second goal we use the method in Section \ref{preference-based-explanations} to compare policies $\pi^{p}_A$ and $\pi^{p}_B$ for every $p \in \begin{bmatrix} 0, 1, 2,  3, 4\end{bmatrix}$. These policies should not differ in their abilities or preferences, since they are trained for same amount of time steps and on the same reward function. Finally, to investigate the third evaluation goal we compare the behavior of $\pi^{p}_{A}$ and $\pi^{p}_{rand}$ for every $p \in \begin{bmatrix} 0, 1, 2, 3, 4\end{bmatrix}$. These two policies differ in their abilities, as $\pi^{p}_{rand}$ has been trained for a shorter time, and fails to learn any meaningful strategy in the environment.  For all three evaluation scenarios we record the total number of disagreements encountered as well as the number of preference-based disagreements according to the method presented in Section \ref{preference-based-explanations} and generate contrasting explanations provided that the number of gathered disagreements exceeds $0$. Further parameters for this approach are given in Table \ref{method-params}.

\begin{table}[t]
    \centering
    \caption{Parameters for data collection and analysis process presented in Section \ref{preference-based-explanations}.}
    \begin{adjustbox}{width=1\linewidth}
        \begin{tabular}{@{}ccc@{}}
            \toprule
                Parameter & Description & Value \\
                \midrule
                $n$ & Number of data-collection episodes & 1000\\
                $k$ & Maximum length of disagreement trajectory & 10\\
                $\alpha$ & State importance threshold & 0.8\\
                $\beta$ & Disagreement state evaluation similarity threshold & 0.1 \\
                $\gamma$ & Disagreement outcome similarity threshold & 0.1  \\
                $p_{thres}$ & p-value threshold & 0.05 \\
                
            \bottomrule
        \end{tabular}
    \end{adjustbox}
    \label{method-params}
\end{table}

\begin{table}[t]
    \centering
    \caption{Total number of disagreements and number of preference-based disagreements obtained by applying method from Section \ref{preference-based-explanations} on the three evaluation scenarios.}
    \begin{adjustbox}{width=1\linewidth}
        \begin{tabular}{@{}cccc@{}}
            \toprule
                Parameter $p$ & Evaluation scenario & \makecell{Total number \\of disagreements} & \makecell[c]{Number of \\preference-based disagreements} \\ \midrule
                1 & \makecell[c]{$\pi_A$ vs $\pi_{safe}$\\
                                 $\pi_A$ vs $\pi_B$\\
                                 $\pi_A$ vs $\pi_{rand}$}
                  & \makecell[c]{1490\\2910\\2691}
                  & \makecell[c]{0\\0\\0} \\ \midrule
                2 & \makecell[c]{$\pi_A$ vs $\pi_{safe}$\\
                                 $\pi_A$ vs $\pi_B$\\
                                 $\pi_A$ vs $\pi_{rand}$}
                  & \makecell[c]{2779\\2000\\2001}
                  & \makecell[c]{182\\0\\0} \\ \midrule
                3 & \makecell[c]{$\pi_A$ vs $\pi_{safe}$\\
                                 $\pi_A$ vs $\pi_B$\\
                                 $\pi_A$ vs $\pi_{rand}$}
                  & \makecell[c]{2464\\1843\\2006}
                  & \makecell[c]{216\\0\\0} \\ \midrule
                4 & \makecell[c]{$\pi_A$ vs $\pi_{safe}$\\
                                 $\pi_A$ vs $\pi_B$\\
                                 $\pi_A$ vs $\pi_{rand}$}
                  & \makecell[c]{2756\\2000\\3000}
                  & \makecell[c]{927\\0\\0} \\
            \bottomrule
        \end{tabular}
    \end{adjustbox}
    \label{num-disagreements}
\end{table}

\begin{figure}[t]
    \centering
    \begin{minipage}{0.45\linewidth}
        \centering
        \includegraphics[width=1\textwidth]{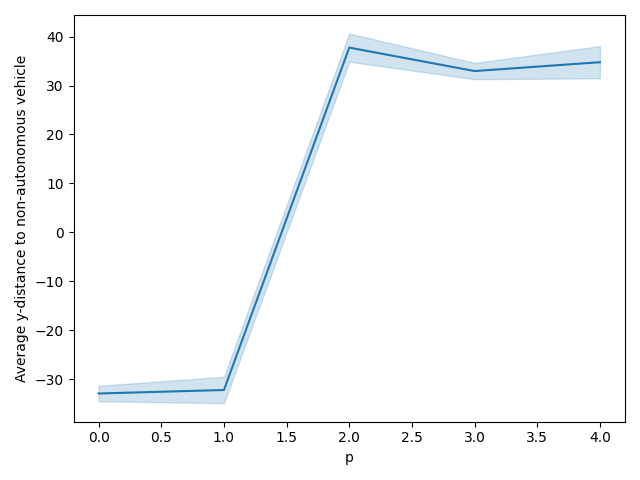}
        \caption{Average y-distance }
        \label{y-dist}
    \end{minipage}\hfill
    \begin{minipage}{0.45\linewidth}
        \centering
        \includegraphics[width=\textwidth]{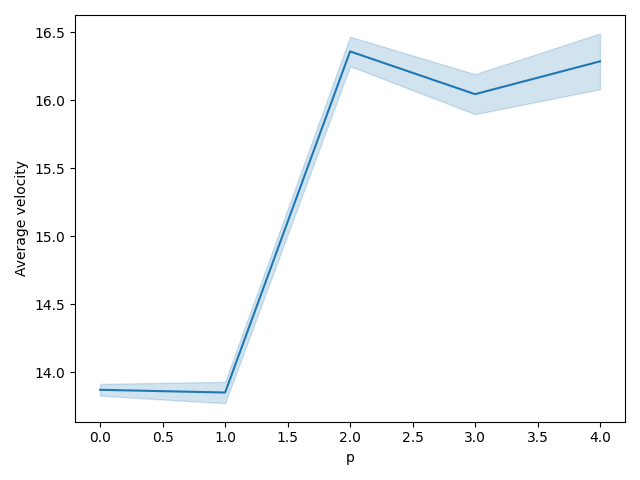}
        \caption{Average velocity throughout episode}
        \label{avg-vel}
    \end{minipage}
    \label{feature-plot}
\end{figure}

Results for the three evaluation scenarios are presented in Table \ref{num-disagreements}. Firstly, we can notice that in each scenario policies have encountered a number of  disagreement states, despite the fact that some policies, such as $\pi^p_A$ and $\pi^p_B$ were trained on the same reward function for the same amount of time. From Table \ref{num-disagreements} we can clearly see that our method abides by the evaluation goals $2$ and $3$, because comparing two capable policies trained with the same preferences ($\pi^p_A$ vs. $\pi^p_B$) or two policies with different abilities ($\pi^p_A$ vs. $\pi^p_B$) yields no preference-based disagreements. In order to confirm that the method also satisfies the first evaluation goal, we must determine whether actual behavior of policies corresponds to number of discovered preference-based disagreements. In other words, according to Table \ref{num-disagreements} the proposed approach concluded that only policy $\pi^1_A$ follows the same safety-oriented strategy as the $\pi_{safe}$ and merges behind the non-autonomous vehicle, while all other policies $\pi^2_A, \pi^3_A, \pi^4_A$ exhibit a different, more aggressive behavior. To verify these findings, we explore the behavior of these policies and for each of them record the average y-coordinate distance at the moment of merging into the right lane (Figure \ref{y-dist}). Similarly, we record the average velocity of the car during the episode for each of the trained policies (Figure \ref{avg-vel}) since this feature is the most indicative of the agent's strategy -- in order to merge in front of the non-autonomous vehicle agent must speed up, and for safety-oriented merging agent will need to slow down. Figures \ref{y-dist} and \ref{avg-vel} show that the behavior of policy $\pi^1_A$ differs significantly from the other policies, and is most similar to $\pi_{safe}$. Specifically, $\pi^1_A$ is the only policy which merges behind the non-autonomous vehicle (Figure \ref{y-dist}) and maintains an average speed lower than the initial velocity (Figure \ref{avg-vel}). From this we can conclude that our method indeed satisfies the first evaluation goal. 

Finally, we use the preference-based disagreement data to generate explanations about contrasting behavior of two policies with different strategies using the method described in Section \ref{generating-exp}. The explanations identify the state feature values that certain agents prefers compared to the other agent. Specifically, after comparing two policies $\pi_A$ and $\pi_B$, where $\pi_A$ is safely-oriented, while $\pi_B$ prefers more aggressive driving, we obtain the following explanations:

\begin{displayquote}

\textit{Policy $\pi_A$ prefers states with $x$ smaller, $y$ smaller, $h$ smaller, $v$ smaller, $u$ larger
compared to policy $\pi_B$}.

\end{displayquote}

\section{Discussion and future work}

In this work we focused on the problem of explaining differences in behavior of two RL agents that stem from their opposing preferences. We proposed a method for distinguishing between ability-based and preference-based differences and generated contrasting explanations about state feature values that agents prefer. We also evaluated our approach on a merging task in autonomous driving. 

Although we have shown that our method can successfully differentiate between ability and preference-based differences in behavior, our approach relies on the choice of threshold values $\alpha$, $\beta$ and $\gamma$. Additionally, our approach focuses on comparing only two RL policies. In future work we hope to address these two limitations, and extend the method to allow for end-to-end approach for learning threshold parameters and to support multiple policies. 

\section*{Acknowledgement}

This publication has emanated from research supported in part by a grant from Science Foundation Ireland under Grant number 18/CRT/6223 . For the purpose of Open Access, the author has applied a CC BY public copyright licence to any Author Accepted Manuscript version arising from this submission.

\bibliography{references}
\end{document}